\documentclass{article}
\usepackage{spconf,amsmath,graphicx}

\usepackage[utf8]{inputenc} 
\usepackage[T1]{fontenc}    
\usepackage{hyperref}       
\usepackage{url}            
\usepackage{booktabs}       
\usepackage{amsfonts}       
\usepackage{nicefrac}       
\usepackage{microtype}      
\usepackage[inline]{enumitem}
\usepackage{subfig}
\usepackage{float}
\usepackage{multirow}
\usepackage{array}
\usepackage{adjustbox}
\usepackage{tabularx}
\usepackage{xspace}

\usepackage{amsmath}
\usepackage{amssymb}
\usepackage{amsthm}
\usepackage{graphicx}
\usepackage{xcolor,colortbl}

\title{Memory-based Message Passing: Decoupling the Message for Propagation from Discrimination}

\name{Jie Chen$^{\star}$ \qquad Weiqi Liu$^{\star}$ \qquad Jian Pu$^{\#}$  \sthanks{Corresponding authour: jianpu@fudan.edu.cn}
\thanks{This paper is supported by National Natural Science Foundation of China (No. 62176059), Shanghai Municipal Science and Technology Major Project (No. 2018SHZDZX01), and ZJ Lab, and National Key Research and Development Program of China (No. 2018YFB1305104).}
}
  
  \address{$^{\star}$ Shanghai Key Lab of Intelligent Information Processing and School of Computer Science,\\ Fudan University, Shanghai, China \\
      $^{\#}$  Institute of Science and Technology for Brain-Inspired Intelligence (ISTBI), \\ Fudan University, Shanghai, China}

\begin{document}

\maketitle
\begin{abstract}
Message passing is a fundamental procedure for graph neural networks in the field of graph representation learning. Based on the homophily assumption, the current message passing always aggregates features of connected nodes, such as the graph Laplacian smoothing process. However, real-world graphs tend to be noisy and/or non-smooth. The homophily assumption does not always hold, leading to sub-optimal results. A revised message passing method needs to maintain each node's discriminative ability when aggregating the message from neighbors. To this end, we propose a Memory-based Message Passing (MMP) method to decouple the message of each node into a self-embedding part for discrimination and a memory part for propagation. 
Furthermore, we develop a control mechanism and a decoupling regularization to control the ratio of absorbing and excluding the message in the memory for each node. 
More importantly, our MMP is a general skill that can work as an additional layer to help improve traditional GNNs performance.
Extensive experiments on various datasets with different homophily ratios demonstrate the effectiveness and robustness of the proposed method.
\end{abstract}

\section{Introduction}
Recently, the emerging graph neural networks (GNNs) have demonstrated powerful abilities in semi-supervised node classification tasks\cite{tnnls-compre}.
Most GNNs follow a message passing architecture~\cite{gilmer2017neurala}. In each GNN layer, each node aggregates information from its neighbors and then updates the node feature. This message passing mechanism underlies the assumption of local homophily~\cite{zhu2020beyond}, i.e. connected nodes tend to be the same class. It can be seen as a Laplacian smoothing progress~\cite{li2018deeper} which smooth each node's embedding by neighbors. Although the local homophily assumption seems reasonable and helpful to achieve good prediction results for homophilous graphs such as citation networks, it also limits GNNs’ applications in heterophilous graphs~\cite{bo2021beyond,zhu2020beyond}.

Heterophilous graphs widely exist in real-world society. For instance, nodes in actor networks with diverse roles are more likely to be connected, and different amino acid types tend to form connections in protein structures~\cite{zhu2020beyond}. In such heterophilous graphs, the performance of classical message passing graph neural networks such as GCN~\cite{GCN} and GAT~\cite{GAT} is even inferior to a simple multi-layer perceptron (MLP) in node classification tasks~\cite{pei2019geom}. 
The reason for such degradation is due to the massive presence of different labels for connected nodes in the heterophilous graph. Message passing among these nodes may bring negative disturbance~\cite{hou2019measuring,chen2021graph} and blur the classification boundary for node classification.
However, determining whether a graph is homophily or not is a challenging task due to the lack of labels~\cite{zhu2021graph}.
Hence, the message passing mechanism needs to be revised to adaptively maintain each node's discriminative ability when aggregating the message from its neighbors. 

Recently, various works have investigated GNN’s ability to improve the message passing scheme. On the one hand, new structures are also emerging to deal with the heterophilous graph~\cite{pei2019geom,bo2021beyond,zhu2020beyond,chien2020adaptive}. Geom-GCN~\cite{pei2019geom} utilizes the structural similarity to capture the smooth structure and long-range dependencies in non-smooth heterophilous graphs. FAGCN~\cite{bo2021beyond} divides the message into low-frequency and high-frequency signals to deal with heterophilous graphs. But they usually use specific architecture and thus are incompatible with traditional GNNs. On the other hand, some plug-in components such as 
Jump Knowledge(JK)~\cite{xu2018representation} introducing the long-range ability, and DropEdge(DE)~\cite{rong2019dropedge} acting as a message passing reducer
can be placed with traditional GNN to improve their performance. However, they usually lack adaptivity since they are not designed for heterophilous graphs. Consequently, a well-generalizable message passing method should perform well on graphs regardless of homophilous property and need to be compatible with traditional GNNs as a plug-in component.

To this end, we propose a novel Memory-based Message Passing (MMP) mechanism which decouples the messages into two parts, i.e., memory for propagation and self-embedding for discrimination. Specifically, each node is endowed with a memory cell and sends messages from the memory cell instead of hidden self-embedding. After propagation, each node can leverage a learnable control mechanism to adaptive update its self-embedding and memory cell according to their recent states. A decoupling regularization loss function is also applied to enhance the disentanglement for feature propagation and discrimination. Moreover, our MMP works as an additional layer that can easily plug into any classic GNN model and improve their performance on heterophilous graphs.  
 
The contribution of this work is summarized as follows: (1) We generalize the message passing of GNN to MMP and show that MMP can provide substantial improvements for heterophilous graphs. (2) We show that MMP is more robust to the noisy graph scenario than the non-memory counterparts. (3) We apply additional decoupling regularization for message passing to enhance the independence of node discrimination and propagation. 

\section{Background}
\subsection{Notations and Problem Setting}
Consider an undirected graph $\cal G = (V, E)$ with adjacency matrix $\mathbf{A}\in R^{N\times N}$, where $\cal V$ is a set of nodes with ${\cal V}=\{v_1, \cdots, v_N\}$ and $\cal E$ is a set of edges. For each node $v_i\in {\cal V}$, we denote $\mathcal{N}(i)=\{j:(i,j)\in {\cal E}\}$ as its neighbor set according to the edge set ${\cal E}$. Each node has a d-dimensional feature representation $\mathbf{x}_i$ and a c-dimensional one-hot class label $\mathbf{y}_i$, where
$\mathbf{X}=[\mathbf{x}_1, \cdots, \mathbf{x}_N]$ and $\mathbf{Y}=[\mathbf{y}_1, \cdots, \mathbf{y}_N]$.

Given the labels ${\bf Y}_{\cal L}$ of the nodes ${\cal L} \subset {\cal V}$, the task of semi-supervised node classification is to predict the labels ${\bf Y}_{\cal U}$ of the unlabeled nodes ${\cal U} = {\cal V} \setminus {\cal L}$ by exploiting the graph structure ${\cal E}$ and the features of nodes $\mathbf{X}$.

\subsection{General Message Passing }
The general message passing is composed of an aggregate step and an update step. Each node first aggregates information from its neighbors then update its self-embedding as follows:
\begin{align}
\mathbf{M}_i^l &= \mathrm{aggregate}(\mathbf{H}_j^{l-1}; v_j\in \mathcal{N}(i)) \\
\mathbf{H}_i^l &= \mathrm{update}(\mathbf{H}_i^{l-1}, \mathbf{M}_i^l).
\end{align}
For the $l$-th layer of a GCN, we use $\mathbf{h}^{l}_i$ to represent the embedding of node $i$, the $\mathbf{h}^{0}_i$ can be set to $\mathbf{x}_i$ or a projection of $\mathbf{x}_i$ for dimension reduction, the $\mathbf{m}^{l}_i$ denotes the aggregating message, and $({\bf W}^{l}$,  $\mathbf{b}^{l})$ to denote the corresponding weights and bias, and $\sigma(\cdot)$ to be the non-linear activation function. The general GCN message passing rule for the $l$-th layer for node $i$ is usually formulated by:
\begin{align}
    \mathbf{m}^{l}_i &= \sum\limits_{j\in\mathcal{N}(i)} v_{ij}\mathbf{h}^{l-1}_j, & \text{(aggregate)}
    \label{eqn::feat_agg} \\ 
    \mathbf{h}^{l}_i &= \sigma( {\mathbf{W}}^{l}\mathbf{h}^{l-1}_i + \mathbf{W}^{l}\mathbf{m}^{l}_i + \mathbf{b}^{l}), & \text{(update)} \label{eqn::feat_transform}
\end{align}
where $v_{ij}$ denotes the weights for aggregation that can be computed by the adjacent matrix ${\bf A}$~\cite{GCN, Hamilton2017InductiveRL, klicpera2018predict} or attention mechanism~\cite{GAT}. 
The final output $\mathbf{Z} \in R^{N \times c}$ of the label prediction is evaluated using a \textit{softmax} classifier to the last layer $\mathbf{H}^L$. The optimization goal is the cross-entropy loss:

\begin{equation}
    \mathcal{L}_{\text {semi}}=-\sum_{i \in \cal L} \sum_{j=1}^{c} \mathbf{Y}_{i j} \ln \mathbf{Z}_{i j}.
\end{equation}

\section{Proposed Method}
The key idea of our method is to decouple the message into two parts: propagation and discrimination.
We first introduce the memory-based message passing schema, which gives the memory cell to each node and propagates the message from the memory cell instead of its self-embedding (in Section \ref{Sec:MMP}). We then describe how to improve the disentanglement of the message for propagation and discrimination by decoupling regularization (in Section\ref{Sec:Decoupling}).

\subsection{Memory-based Message Passing}\label{Sec:MMP}
The goal of an adaptive message passing mechanism for different graphs is to 
maintain each node's discriminative ability when aggregating the message from its neighbors. 
Inspired by the router in the real-world web net, the message from each node is stored in the memory buffer of the router, and each computer can fetch useful information from the buffer. We argue that the message passing of GNNs should also decouple the message into the feature for propagation and the feature for downstream task discrimination.

To maintain each node's discriminative ability, we propose a memory cell $\mathbf{C}$ for message passing. When performing the aggregate step, each node sends the feature from the memory cell to get message $\mathbf{M}$. In the update step, each node both updates the memory cell and hidden feature ${\bf H}$ for the downstream task based on the previous hidden $\mathbf{H}$ and the received message $\mathbf{M}$. Such procedure allows each node to adaptive choosing whether to absorb or discard the message from neighbors.
\begin{align} 
\mathbf{M}_i^l &= \mathrm{aggregate}(\mathbf{C}_j^{l-1}; v_j\in \mathcal{N}(i))\\
\mathbf{H}_i^l, \mathbf{C}_i^l &= \mathrm{update}(\mathbf{H}_i^{l-1}, \mathbf{M}_i^l)
\end{align}
As mentioned, we can use any other classical graph convolution to aggregate the message from neighbors. The major difference to the traditional message passing mechanism is that we propagate the message from memory cell $\mathbf{C}$ instead of hidden feature ${\bf H}$, and thus there is no need for modification for the classical graph convolution at the code level. Also, we adopt an additional layer to evaluate how to update the hidden $\mathbf{H}$ and memory $\mathbf{C}$.
Therefore, our approach is easily applied to the traditional graph convolution as a plug-in component to improve the performance of heterophilous graphs.

For the update step, we apply a control mechanism $f_\theta$ to calculate the coefficients $\alpha$s to the ratio of absorbing the message and the ratio of discarding the message. The function $f_\theta: \mathbb{R}^{n \times 2d} \rightarrow \mathbb{R}^{n \times 3}$ is set to be a share weight fully connected linear layer with a sigmoid function, so that all the coefficients $\alpha$s are scalar ranging from $\left [ 0,1 \right ]$. We also note that the recurrent model like GRU~\cite{cho2014learning} can also be utilized to implement the $f_\theta$, which we leave for future work. Compared with the vectorized coefficient for controlling each dimension, this scalar $\alpha$s can simplify the optimization process and prevent overfitting. 
\begin{align}
\alpha_h, \alpha_m, \alpha_c &= f_\theta(\mathbf{H}^{l-1}_i,\mathbf{M}^l_i) \\
\mathbf{H}^{l}_i &= \mathbf{H}^{l-1}_i * \alpha_h + \mathbf{M}^l_i * \alpha_m \\ 
\mathbf{C}^l_i &= \mathbf{M}^l_i * \alpha_c
\end{align}

When updating the hidden state $\mathbf{H}^l$, we linearly combine the previous hidden $\mathbf{H}^{l-1}$ and the aggregating message $\mathbf{M}^l$ according to coefficients $\alpha$s. Such an operator may help each node maintain the discriminative ability and aggregate messages adaptively. For instance, when the message from other class neighbors is inconsistent, the $\alpha_m$ can be learned to hit 0 to avoid the update. It is worth noting that the proposed MMP degenerates into traditional message passing by setting $\alpha_h=\alpha_m=0$ and $\alpha_c=1$.

When updating the memory cell $\mathbf{C}$, different from the traditional graph convolution that each node always sends its current embedding, we can clear the node memory by learning $\alpha_m$ close to 0 if the node has no useful information to its neighbors. 
Note that, in this step, we can also use $\mathbf{H}$ to update $\mathbf{C}$, but we found the empirical performances are similar.

\subsection{Decoupling Regularization}\label{Sec:Decoupling}
As mentioned before, the key idea of MMP is to decouple the feature for message passing and discrimination of each node. Since we use the additive model to mixture the embedding, the memory cell and hidden state may be intertwined to increase the difficulty of the decoupling. Inspired by ensemble learning that encourages the base learner to be diverse and independent to achieve better predictions~\cite{krogh1994neural}, we adopt a decoupling regularization term to enhance the disentanglement between the memory cell and hidden state. The overall memory decoupling regularization is formulated as follows:

\begin{align}
\mathcal{L}_{\text {decouple}}=\sum_{i\in V} \sum_{l=0}^L \frac{\left|\left\langle \mathbf{C}_{i}^{l},  \mathbf{H}_{i}^{l}\right\rangle\right|}{\left\|\mathbf{C}_{i}^{l}\right\| \cdot\left\|\mathbf{H}_{i}^{l}\right\|}.
\end{align}

By defining the decoupling loss with the cosine similarity, our approach encourages the increments of the memory and hidden to be orthogonal. We use it to diversify the feature between propagation and discrimination for each node at every layer, maintaining more information during message passing.
We use the hyper-parameter $\lambda$ as the regularization parameter. The final MMP model is trained end-to-end based on the following objective function:
\begin{align}
    \mathcal{L}_{\text {final}} = \mathcal{L}_{\text {semi}} + \lambda \mathcal{L}_{\text {decouple}}.
\end{align}

\section{Experimental Results}

\setlength{\tabcolsep}{3pt}
\begin{table*}%
   \scriptsize
    \centering  
    \vspace{-0.35cm}
    \caption{%
    Real data: mean accuracy $\pm$ stdev over 10 random data splits. The lower homophily ratio h means the higher heterophily(h<0.5). Best model per benchmark highlighted in blue. The "*" results are obtained from~\cite{pei2019geom,zhu2020beyond}. %
    }
    \label{tab:5-real-results}
    \begin{adjustbox}{width=\textwidth}
    \begin{tabular}{lcccccccccc} %
    \toprule
       &  \texttt{\bf Texas}           &   \texttt{\bf Wisconsin}           &   \texttt{\bf Actor}            &   \texttt{\bf Squirrel}   &   
       \texttt{\bf Chameleon} &
       \texttt{\bf Cornell}    &   
       \texttt{\bf Citeseer}           &   \texttt{\bf Pubmed}            &   \texttt{\bf Cora} &  
            {\texttt{\bf Avg Rank}}
       \\
          \textbf{Hom.\ ratio} $h$ & \textbf{0.11} & \textbf{0.21} & \textbf{0.22} & \textbf{0.22} & \textbf{0.23} & \textbf{0.3} & 
          \textbf{0.74} & \textbf{0.8} & \textbf{0.81} & 
          -
          \\
    \midrule
    {GEOM-GCN*\cite{pei2019geom}} & $67.57$ & $64.12$ & $31.63$ & $38.14$ & $60.90$ & $60.81$ 
	   & \cellcolor{blue!15}$\bf{77.99}$ &  \cellcolor{blue!15}$\bf{90.05}$ & $85.27$ & 6.7 \\

	   {MixHop*}\cite{abu2019mixhop} & $77.84{\scriptstyle\pm7.73}$ & $75.88{\scriptstyle\pm4.90}$ & $32.22{\scriptstyle\pm2.34}$ & $43.80{\scriptstyle\pm1.48}$ & $60.50{\scriptstyle\pm2.53}$ & $73.51{\scriptstyle\pm6.34}$ & 
	   $76.26{\scriptstyle\pm1.33}$ & $85.31{\scriptstyle\pm0.61}$ & \cellcolor{blue!15}$\bf{87.61}{\scriptstyle\pm0.85}$ & 6.0 \\
       {H2GCN*\cite{zhu2020beyond}} & $84.86{\scriptstyle\pm6.77}$ & \cellcolor{blue!15}$\bf{86.67}{\scriptstyle\pm4.69}$ & $35.86{\scriptstyle\pm1.03}$ & $36.42{\scriptstyle\pm1.89}$ & $57.11{\scriptstyle\pm1.58}$ & $82.16{\scriptstyle\pm4.80}$ & 
       $77.07{\scriptstyle\pm1.64}$ & $89.40{\scriptstyle\pm0.34}$ & $86.92{\scriptstyle\pm1.37}$ & 4.1\\
	   {GPRGNN\cite{chien2020adaptive}} & $78.64{\scriptstyle\pm6.42}$ & $84.31{\scriptstyle\pm5.54}$ & $34.63{\scriptstyle\pm1.11}$ & $41.78{\scriptstyle\pm1.63}$ & $62.85{\scriptstyle\pm1.68}$ &  $77.83{\scriptstyle\pm5.47}$ & 
	   $77.26{\scriptstyle\pm2.09}$ & 
	   $88.25{\scriptstyle\pm0.72}$ & 
	   $87.60{\scriptstyle\pm1.31}$ & 4.3 \\ 
	   {FAGCN\cite{bo2021beyond}} & $77.56{\scriptstyle\pm6.11}$ & $79.41{\scriptstyle\pm6.55}$ & $34.85{\scriptstyle\pm1.61}$ & $30.59{\scriptstyle\pm1.22}$ & $46.44{\scriptstyle\pm2.81}$ & $78.64{\scriptstyle\pm5.47}$ & 
	   $74.01{\scriptstyle\pm1.85}$ & 
	   $76.57{\scriptstyle\pm1.88}$ & 
	   $86.34{\scriptstyle\pm0.67}$ & 9.0\\ 
	   {MLP*} & $81.89{\scriptstyle\pm4.78}$ & $85.29{\scriptstyle\pm3.61}$ & $35.76{\scriptstyle\pm0.98}$ & $29.68{\scriptstyle\pm1.81}$ & $46.36{\scriptstyle\pm2.52}$ & $81.08{\scriptstyle\pm6.37}$ & 
	   $72.41{\scriptstyle\pm2.18}$ & $86.65{\scriptstyle\pm0.35}$ & $74.75{\scriptstyle\pm2.22}$ & 8.6 \\
       \midrule

	   {GAT*\cite{GAT}} & $58.38{\scriptstyle\pm4.45}$ & $55.29{\scriptstyle\pm8.71}$ & $26.28{\scriptstyle\pm1.73}$ & $30.62{\scriptstyle\pm2.11}$ & $54.69{\scriptstyle\pm1.95}$ & $58.92{\scriptstyle\pm3.32}$ & 
	   $75.46{\scriptstyle\pm1.72}$ & $84.68{\scriptstyle\pm0.44}$ & $82.68{\scriptstyle\pm1.80}$ & 12.0 \\
	   {GAT+JK} & $64.59{\scriptstyle\pm6.74}$ & $59.80{\scriptstyle\pm7.52}$ & $31.11{\scriptstyle\pm1.18}$ & $35.41{\scriptstyle\pm1.97}$ & $57.61{\scriptstyle\pm3.89}$ & $58.11{\scriptstyle\pm8.94}$ & 
	   $75.47{\scriptstyle\pm1.30}$ & $86.38{\scriptstyle\pm0.32}$ & $87.21{\scriptstyle\pm1.40}$ & 9.4 \\ 
        {GAT+DE} & $57.02{\scriptstyle\pm6.42}$ & $54.31{\scriptstyle\pm7.39}$ & $28.91{\scriptstyle\pm1.21}$ & $36.50{\scriptstyle\pm2.23}$ & $56.72{\scriptstyle\pm3.34}$ & $60.81{\scriptstyle\pm8.37}$ & 
	   $75.99{\scriptstyle\pm1.58}$ & $84.55{\scriptstyle\pm0.38}$ & $86.24{\scriptstyle\pm0.80}$ & 10.7 \\  
        {GAT+MMP(ours)}& $82.54{\scriptstyle\pm5.22}$ & $85.29{\scriptstyle\pm5.16}$ & $36.36{\scriptstyle\pm1.38}$ & $44.32{\scriptstyle\pm2.31}$ & $64.69{\scriptstyle\pm2.75}$ & $80.54{\scriptstyle\pm5.66}$ & 
	   $72.22{\scriptstyle\pm1.46}$ & $82.91{\scriptstyle\pm2.96}$ & $86.21{\scriptstyle\pm0.61}$ & 5.8 \\
       \midrule
	   {GCN*\cite{GCN}} & $59.46{\scriptstyle\pm5.25}$ & $59.80{\scriptstyle\pm6.99}$ & $30.26{\scriptstyle\pm0.79}$ & $36.89{\scriptstyle\pm1.34}$ & $59.82{\scriptstyle\pm2.58}$ & $57.03{\scriptstyle\pm4.67}$ & 
	   $76.68{\scriptstyle\pm1.64}$ & $87.38{\scriptstyle\pm0.66}$ & $87.28{\scriptstyle\pm1.26}$ & 9.7 \\
        {GCN+JK*\cite{xu2018representation}} & $66.49{\scriptstyle\pm6.64}$ & $74.31{\scriptstyle\pm6.43}$ & $34.18{\scriptstyle\pm0.85}$ & $40.45{\scriptstyle\pm1.61}$ & $63.42{\scriptstyle\pm2.00}$ & $64.59{\scriptstyle\pm8.68}$ & 
        $74.51{\scriptstyle\pm1.75}$ & $88.21{\scriptstyle\pm0.45}$ & $85.79{\scriptstyle\pm0.92}$ & 7.4\\
        {GCN+DE\cite{rong2019dropedge}} & $52.71{\scriptstyle\pm5.13}$ & $50.19{\scriptstyle\pm6.68}$ & $28.86{\scriptstyle\pm1.15}$ & $36.83{\scriptstyle\pm2.16}$ & $59.93{\scriptstyle\pm2.78}$ & $61.89{\scriptstyle\pm8.77}$ & 
        $76.81{\scriptstyle\pm1.47}$ & $88.32{\scriptstyle\pm0.55}$ & $86.71{\scriptstyle\pm1.38}$ & 8.2\\
	   {GCN+MMP(ours)} & \cellcolor{blue!15}$\bf{85.39}{\scriptstyle\pm3.58}$ & $85.51{\scriptstyle\pm4.61}$ & \cellcolor{blue!15}$\bf{36.69}{\scriptstyle\pm1.38}$ & \cellcolor{blue!15}$\bf{57.38}{\scriptstyle\pm2.41}$ & \cellcolor{blue!15}$\bf{70.08}{\scriptstyle\pm2.12}$ & 
	   \cellcolor{blue!15}$\bf{82.89}{\scriptstyle\pm5.98}$ &
	   $75.31{\scriptstyle\pm1.87}$ & 
	   $88.39{\scriptstyle\pm0.41}$ & 
	   $86.48{\scriptstyle\pm1.18}$ &  \cellcolor{blue!15}\bf{2.9}\\
	   \bottomrule
    \end{tabular}
    \end{adjustbox}
    \vspace{-0.5cm}
\end{table*}

\subsection{Experimental Setup}

\vspace{4pt}
\noindent
\textbf{Datasets.} We evaluate our model performance and recent GNNs on a variety of real-world datasets with edge homophily ratio h\footnote{homophily ratio h is the fraction of edges in a graph which connect nodes that have the same class label.} ranging from strong heterophily to strong homophily. For all benchmarks , we use the feature vectors, class labels, and 10 random splits (48\%/32\%/20\% of nodes per class for train/validation/test) provided in literature~\cite{pei2019geom,zhu2020beyond}.

\vspace{4pt}
\noindent
 \textbf{Baselines.} For baselines we use \textbf{(1)} traditional message passing GNNs:GCN~\cite{GCN} and GAT~\cite{GAT}; \textbf{(2)} recent specific structure tackling heterophily: Geom-GCN\cite{pei2019geom}, H2GCN~\cite{zhu2020beyond}, GPRGNN~\cite{chien2020adaptive}, FAGCN~\cite{bo2021beyond}, MixHop~\cite{abu2019mixhop}; \textbf{(3)} the plug-in component that improve the message passing performance: JK~\cite{xu2018representation}, DropEdge(DE)~\cite{rong2019dropedge}; \textbf{(4)} standard 2-layer MLP. For ease of comparison, we use the reported results in literature~\cite{pei2019geom,zhu2020beyond}. Moreover, for the missing results, we rerun their released code over 10 times.
 
\vspace{4pt}
\noindent
\textbf{Implementation Details.} For MMP, we use 2 hidden layers, 64 hidden dimensions, 
0.5 dropout ratio, and Adam optimization method~\cite{kingma2014adam}
to train our model for 500 epochs with early stop. During training, we set the learning rate as 0.05 and weight decay as 0.0005 for all datasets. For the hyper-parameter $\lambda$, we select it in ${\{0, 0.1, 0.2, 0.4, 0.6, 0.8, 1\}}$ using the validation set.

\subsection{Standard Node Classification} 
Table~\ref{tab:5-real-results} summarizes the classification performance of MMP and the baseline methods on all datasets. Note that the result of our MMP with traditional GNN can achieve a comparable result with the state-of-the-art GNN designed for heterophilous datasets. Moreover, compared with the basic GNN models and the message passing improvement methods(JK~\cite{xu2018representation}, DE~\cite{rong2019dropedge}), the MMP can consistently achieve superior performance on the heterophilous datasets and comparable performance on homophilous datasets. It is worth noting that the GCN+MMP achieves the highest avg ranking over all datasets.
This demonstrates the importance of decoupling the message for propagation from downstream task discrimination. 

\subsection{Robustness of MMP}
To evaluate the robustness of the memory based message passing on noisy graphs, we construct graphs with random edge additions following the literature~\cite{NEURIPS2020_e05c7ba4}. Specifically, we randomly add 25\%, 50\%, 75\%, 100\%, 200\%, 300\%, 400\% and 500\% of the edges in the original graphs. 

\begin{figure}[tb]
\begin{minipage}[b]{1.0\linewidth}
  \centering
  \centerline{\includegraphics[width=9cm]{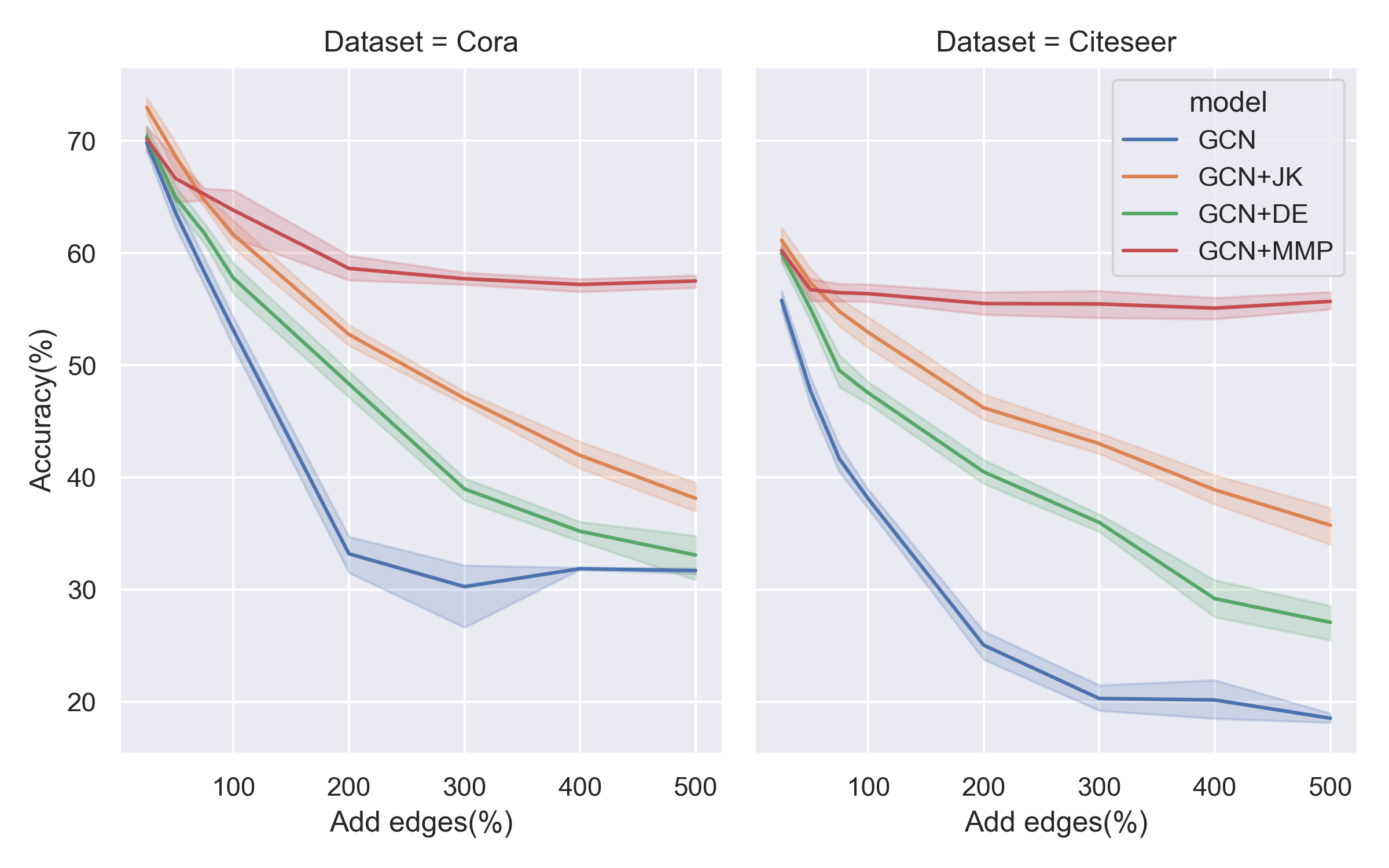}}
\end{minipage}
\caption{Test accuracy($\pm$ std) in percentage for the edge addition scenario on Cora and Citeseer.}
\label{fig:res}
\end{figure}

As shown in Fig.~\ref{fig:res}, though the prediction performance is slightly affected in the beginning, our MMP achieves significant better prediction accuracies for highly noisy graphs (Add edges>200\%).
The performance comparison clearly proves that the memory and decoupling regularization can effectively alleviate the negative disturbance from the noisy message sent by randomly connected neighbors.

\begin{figure}[tb]
\centering
\begin{minipage}[b]{0.7\linewidth}
  \centering
  \centerline{\includegraphics[width=6.5cm]{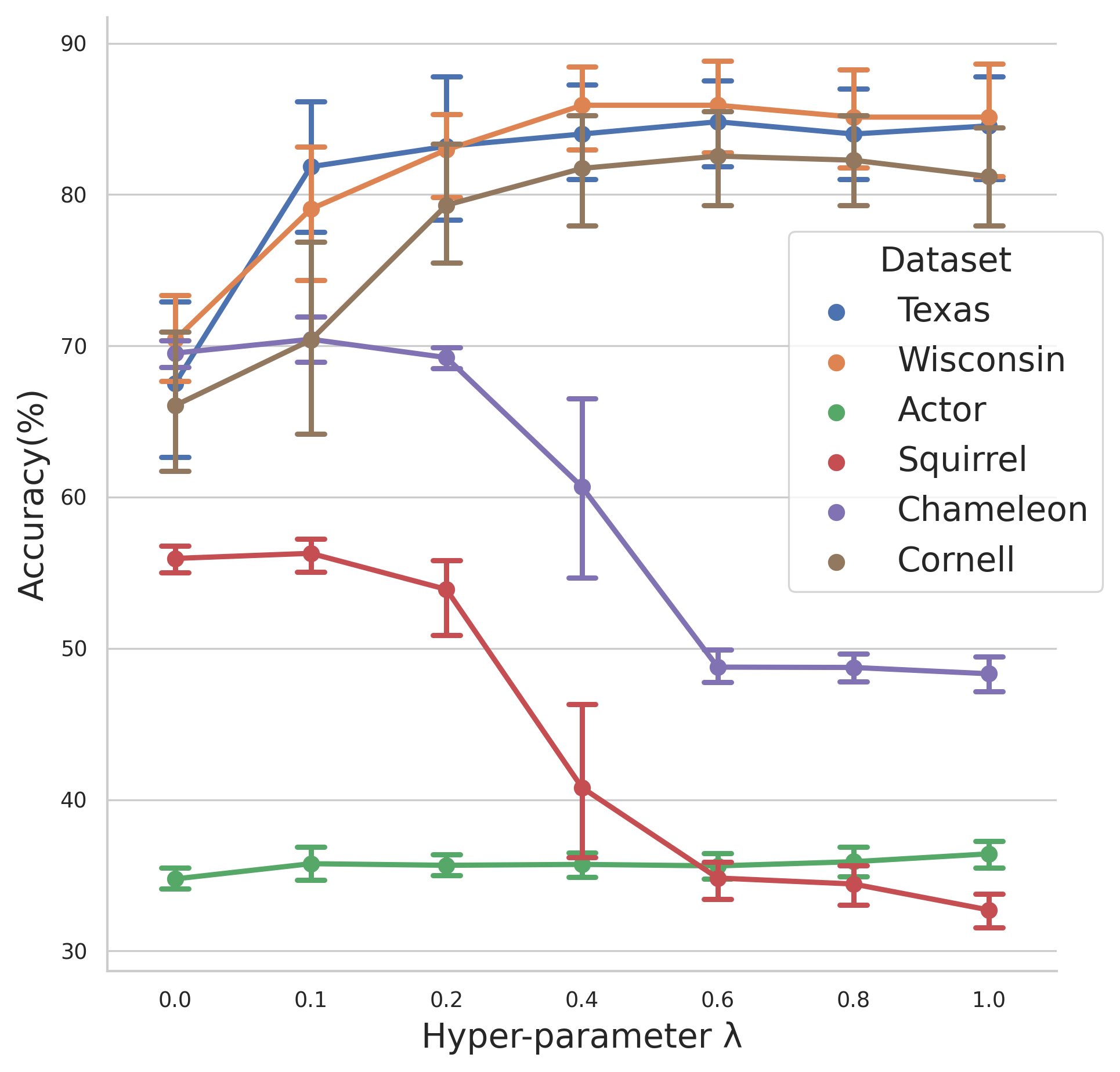}}
\end{minipage}
\caption{Prediction accuracy versus hyper-parameter $\lambda$.}
\label{fig:hyper}
\end{figure}

\subsection{Sensitivity to Hyper-parameters}
We also analysis the prediction sensitivity of parameter $\lambda$ on the heterophilous datasets in Fig.~\ref{fig:hyper}. 
When $\lambda$ is set to 0, there is no decoupling regularization for MMP. Notice that for three datasets (Texas, Wisconsin, Cornell), the decoupling regularization is extremely useful when $\lambda$ is increasing. For the Actor dataset, the regularization is slightly helpful.
However, when $\lambda$ is too large, 
the regularization may force MMP to discard relevant message information in feature aggregation,
which make the performance degenerate for Squirrel and Chameleon.

\section{Conclusion}
In this paper, we have proposed a Memory-based Message Passing (MMP) to decouple the message for propagation from the discrimination by sending the node memory instead of its self-embedding. We develop a learnable control mechanism for each node to help determine the ratio of absorbing and discarding the message. We also propose a decoupling regularization to help each node diversify the message at every layer during message passing. Moreover, our MMP can work as an additional layer that easily plugs into traditional GNNs to improve their performance. Extensive experiments show our model achieves the state-of-the-art on the real-world homophilous, heterophilous, and noisy synthesis graph datasets.

\vfill\pagebreak
\bibliographystyle{IEEEbib}
\bibliography{strings,refs}
\end{document}